\documentclass{article}

\usepackage[utf8]{inputenc} 
\usepackage[T1]{fontenc}    
\usepackage{hyperref}       
\usepackage{url}            
\usepackage{booktabs}       
\usepackage{amsfonts}       
\usepackage{nicefrac}       
\usepackage{microtype}      
\usepackage[dvipsnames]{xcolor}         
\usepackage{amsmath}
\usepackage{multirow}
\usepackage{makecell}
\usepackage{graphicx}

\usepackage{lipsum}
\usepackage{geometry}
\usepackage{authblk}
\usepackage{verbatim}
\usepackage{amssymb}
\usepackage{amsfonts}
\usepackage{array, etoolbox}
\usepackage{longtable}
\usepackage{multirow}
\usepackage{multicol}
\usepackage{caption}
\usepackage{subcaption}
\usepackage{adjustbox}
\usepackage{url}
\usepackage{accents}
\usepackage{dsfont}
\usepackage{bbm}
\usepackage{subfiles}
\usepackage{pdflscape}

\hypersetup{
    colorlinks=true,
    linkcolor=blue,
    filecolor=magenta,      
    urlcolor=cyan,
}
\graphicspath{{images/}{../images/}}

\title{Learning Generalized Causal Structure in Time-series}

\author{%
Aditi Kathpalia\vspace{-2ex}\\ Department of Complex Systems, Institute of Computer Science of the Czech Academy of Sciences \\
  Prague, Czech Republic \\
  \texttt{kathpalia@cs.cas.cz} \\
  \and
  Keerti Panchakshari Charantimath\vspace{-2ex} \\
  Department of Mathematics, Indian Institute of Technology, Kharagpur \\
  West Bengal, India \\
  \texttt{kay\_see101@iitkgp.ac.in} \\
  \and
  Nithin Nagaraj\vspace{-2ex} \\
  Consciousness Studies Programme, National Institute of Advanced Studies\\
  Indian Institute of Science Campus, Bengaluru, India\\
  \texttt{nithin@nias.res.in}
}

\begin{document}

\maketitle
\begin{abstract}

The science of causality explains/determines `cause-effect' relationship between the entities of a system by providing mathematical tools for the purpose. In spite of all the success and widespread applications of machine-learning (ML) algorithms, these algorithms are based on statistical learning alone. Currently, they are nowhere close to 'human-like' intelligence as they fail to answer and learn based on the important ``Why?'' questions. Hence, researchers are attempting to integrate ML with the science of causality. Among the many causal learning issues encountered by ML, one is that these algorithms are dumb to the temporal order or structure in data. In this work we develop a machine learning pipeline based on a recently proposed `neurochaos' feature learning technique (\emph{ChaosFEX} feature extractor), that helps us to learn generalized causal-structure in given time-series data. 
\end{abstract}

\section{Introduction}

Current machine learning (ML) algorithms are based on statistical learning, that is, these algorithms learn mostly by identifying associations in the given data. As the popularity and range of ML algorithms have increased to deal with specific tasks of classification and prediction, researchers focus on the task of bringing these ML algorithms closer to human-like intelligence so that these algorithms can be used to deal with \emph{higher-level} problems. One of the major ways in which human intelligence differs from machine intelligence is that humans learn through `cause-effect' reasoning. Put more specifically, humans have the capability to answer \emph{`what if'} kind of questions. \emph{What if I intervene and do something to a given system? What if I had acted in a different way?} Machine intelligence is still far away from answering these kind of questions~\cite{pearl2018book, scholkopf2019causality}.

Let alone learning based on a causal understanding, machines are not even capable of making sense of temporal or sequential order in the data. Further, they lack the ability of \emph{generalization} which involves transfer of learning from one problem to another and learning features that categorize together datasets that are more alike than different. This lack of generalization stems mainly from an inability of these algorithms to learn causal structures~\cite{scholkopf2019causality}.

While a lot of research is being currently pursued at the intersection of causality and machine learning, we focus on causal learning based on time series data. Time series data are real valued data points arranged in a chronological order. Such data is often collected as measurements in different fields such as finance, medicine, climate science etc. Currently, there are two main tasks for which causal inference is done using time series data. One is studying the effect of treatment or interventions on certain variable of a system. This treatment may be provided as a discrete event or may be continuously given over time. This task is generally referred to as \emph{causal treatment effect estimation} and finds applications in estimating the effects of policies implemented so as to increase/ reduce the sale of certain goods or in estimating the effect of drugs given to patients~\cite{moodie2007demystifying, athey2017state, moraffah2021causal}. The second one is discovering causal relationships between different variables of a system, for which temporal evolution data is available. This task is generally referred to as \emph{causal network discovery} and is useful in domains where we wish to study the interaction between different variables such as the role of human and natural factors in climate change~\cite{leng2019reconstructing, huang2020detecting, moraffah2021causal}.

For our work, we take one step backward and ask the question of whether ML algorithms can even identify whether given time series has an inherent causal structure associated with it, that is, whether the past values of a time series effect its present values or if that is not the case. Learning this structure for time series is essential for making sense of any ordered data given as input to ML algorithms and is also essential to meaningfully answer the two causality tasks discussed above. 
There are of course mathematical techniques that rely on fitting autoregressive models and testing for time dependence by estimating serial correlation in given time series. The latter techniques include the Durbin–Watson test~\cite{durbin1950testing}, Ljung–Box test~\cite{ljung1978measure} and Breusch–Godfrey test~\cite{breusch1978testing}. However, these methods have their limitations: they are parametric, requiring the choice of maximum order; do not work with overlapping data or in the case of heteroskedasticity in the error process. Further, since they are purely statistical, based on autocorrelation, they cannot handle data with intervention(s)/ perturbation(s) where the cause-effect relation in time series builds up as a result of external event(s). We are interested in developing a non-parametric way of identifying causal structure in time series data that works irrespective of the underlying model and learns \emph{generalized} causal-structure in data, unaffected by distribution or model shifts. For this task, we rely on extracting strong features which can identify whether or not there is an underlying causal-structure and later using a simple ML algorithm, classify given time-series data based on having a causal-structure or not. We have used time series values directly, frequency domain characteristics of the process, and `neurochaos' based features extracted from the frequency representation of the process to train the ML classifiers in separate models. \emph{Neurochaos} inspired feature learning~\cite{harikrishnan2020neurochaos} has been proposed based on a \emph{ChaosNet} neural network architecture~\cite{balakrishnan2019chaosnet} that is composed of neurons, which individually mimic chaotic firing properties of the human brain neurons. We compare the performance of these models and show that neurochaos based learning is able to extract meaningful features for learning generalized causal-structure in time series data.

This paper is organized as follows: Section~\ref{sec_data} describes the time series data simulated for the study. Section~\ref{sec_methods} describes the methods, that is the features and the classifiers used for distinguishing between causal and non-causal structure in temporal data. Results are given in Section~\ref{sec_res} and we conclude with Section~\ref{sec_discussion}, discussing the results, and hinting at future research directions.

\section{Datasets}
\label{sec_data}
The following time-series datasets were simulated and used in this work:
Time-series having a causal-structure or {\bf causal time-series} were generated using the following models:

\begin{itemize}
    \item {\bf Autoregressive (AR) processes}
    
    AR processes are random processes in which value of the time-series at any time depends linearly on its past values and on a noise term. The general equation governing an AR ($p$) process, $X$, is given as:
\begin{equation}
    X(t)=c+\sum_{i=1}^{p}{a_i X(t-i)}+\varepsilon_t,
    \label{Ar_eq}
\end{equation} 

where, $t$ denotes time, $c$ is a constant, $p$ is the order of the AR process, $a_i$ is the coefficient at lag $i$ time step(s) and $\varepsilon_t$ is the noise term at time $t$. Order of an AR process is the maximum past lag on which the value of the process at any time depends.

AR processes are used to model several processes occurring in nature, economics etc.~\cite{shumway2000time, von2001statistical}

\item {\bf Autoregressive Moving Average (ARMA) processes}

These processes have an AR part and an MA part. Based on the AR part, each term in the time-series is regressed on its past values and based on the MA part, the noise term at any time point is modelled as a linear combination of instantaneous noise term (at that time point) and noise terms at past time points. In mathematical form, an ARMA($p,q$) process $X$ can be expressed as:

\begin{equation}
    X(t)=c+\sum_{i=1}^{p}{a_i X(t-i)}+\sum_{i=0}^{q}{b_i \varepsilon(t-i)},
    \label{Arma_eq}
\end{equation}

where, $t$ denotes time, $c$ is a constant, $p$ is the order of the AR part and $q$ is the order of the MA part, $a_i$ is the AR coefficient at lag $i$ time step(s) and $b_i$ is the MA coefficient at lag $i$ time step(s), $\varepsilon(t)$ is the noise term at time $t$.

ARMA processes are widely used in the modelling of financial and geological processes~\cite{box2015time, shumway2000time}.

\item {\bf Autoregressive Fractionally Integrated Moving Average (ARFIMA)} 
These processes are used to model long term memory processes. An ARMA($p',q$) process when expressed in terms of the lag or backshift operator can be written as:
\begin{equation}
    (1-\sum_{i=1}^{p'}{a_i B^i})X(t)=(1+\sum_{i=1}^{q}{b_i}B^i) \varepsilon(t),
    \label{Arma_lag_eq}
\end{equation}

Now, if the polynomial $(1-\sum_{i=1}^{p'}{a_i B^i})$ has a unit root of multiplicity $d$, then:
\begin{equation}
    (1-\sum_{i=1}^{p'}{a_i B^i})=(1-\sum_{i=1}^{p'-d}{a_i B^i})(1-B)^d,
    \label{Arima_eq}
\end{equation}

This is the property expressed by Autoregressive Integrated Moving Average or ARIMA($p,d,q$) processes with $p=p'-d$. For the case of ARFIMA processes, the difference parameter $d$ is allowed to take non-integer values. They are thus generally expressed as:

\begin{equation}
    (1-\sum_{i=1}^{p'}{a_i B^i})(1-B)^d X(t)=(1+\sum_{i=1}^{q}{b_i}B^i) \varepsilon(t),
    \label{Arfima_eq}
\end{equation}

ARFIMA models are also widely used, for example in the modelling of economic, geological and physiological time series~\cite{granger1980introduction, graves2017brief}.

\end{itemize}

Time-series without a causal structure or {\bf non-causal time-series} were generated as having time-series value at each time point, as independently and randomly chosen from a normal distribution, $\mathcal{N}(\mu,\sigma^2)$ (where $\mu$ and $\sigma$ denote the mean and standard deviation of the distribution) or uniform distribution, $U(b_u,b_l)$ (where $b_u$ and $b_l$ denote the upper bound and lower bound of the distribution).

\section{Methods}
\label{sec_methods}
\subsection{Using time-series values directly}

In this case, time-series values were directly passed through ML classifiers to check if they could classify time series with a causal structure and without one appropriately. For this purpose, Logistic Regression (LR)~\cite{mccullagh2019generalized} and Long-short term memory (LSTM)~\cite{hochreiter1997long} classifiers were used. LSTM classifiers have a recurrent neural network architecture and include not only feedforward but also feedback connections. Along with being applied to single data points, they have also been shown to work for large sequential datasets such as speech and text data. 

\subsection{Using frequency domain characteristics of time-series}

Any periodic structure in a time series is reflected in its frequency domain amplitude. Since continuous causal effect within a time series (where its past values affect present values), can be thought of as imbuing some periodic nature to the time series, we consider it better to analyze the signals in frequency domain. The fast fourier transform algorithm was used to obtain the discrete fourier transform of simulated signals. Frequency domain amplitudes of the signal were then directly passed through ML classifiers: LR and LSTM. 

\subsection{Using ChaosFEX features of the frequency domain signal}

\emph{ChaosNet} is an artificial neural network inspired by the chaotic firing of neurons in the human brain~\cite{balakrishnan2019chaosnet}. Based on ChaosNet, a hybrid learning architecture was proposed later that uses `neurochaos' or `ChaosFEX'~\footnote{The code for ChaosFEX feature extraction is available at https://github.com/pranaysy/ChaosFEX and this code was used in our work. This code is available open source under the license: Apache License, Version 2.0. Consent of the authors was taken to use the code.} features combined with traditional machine learning approaches for the purpose of classification~\cite{harikrishnan2020neurochaos}. The basis of ChaosNet or the extraction of ChaosFEX features is the \emph{topological transitivity} property of Chaos. The ChaosNet architecture is composed of 1-D chaotic neurons called as Generalized Lur{\"o}th Series (GLS) maps. The number of neurons in the architecture depend upon the number of inputs/features provided to the architecture, with each neuron receiving a single value as input for each data instance. These neurons are set to have an initial activity of $q$ units. Input data is first normalized (to lie between $0$ and $1$) and then passed to these GLS neurons. Each GLS neuron starts firing and keeps on firing until it reaches the epsilon neighborhood of the input value (also called stimulus) provided to it. The epsilon neighborhood of a stimulus $y$ is defined as the interval $(y-\varepsilon, y+\varepsilon)$, where $\varepsilon>0$. 

In this work, we use a GLS neuron $T:[0,1) \rightarrow [0,1)$ defined by the following equation:
\begin{equation}
    T(y)=\begin{cases} 
    \frac{y}{b}, &  0 \leq y < b, \\
    \frac{1-y}{1-b}, &  b \leq y < 1,
    \end{cases}
    \label{GLS_neuron}
\end{equation}

where, $y \in [0,1)$ and $0<b<1$. Let the trajectory of a GLS neuron for a stimulus $y$ be given by $A=[q \rightarrow T(q) \rightarrow T^2(q) \ldots \rightarrow T^N(q)]$. Thus, it can be seen that the neuron takes $N$ time steps to reach the epsilon neighborhood of the stimulus. $N$ is referred to as the firing time. The fraction of the time when the trajectory is above the discrimination threshold ($b$) is referred to as the \emph{Topological Transitivity-Symbolic Sequence (TTSS) Feature} for the stimulus $y$.

To classify time-series as having a causal structure or not, normalized frequency domain amplitudes of given time series were passed as input to the ChaosFEX feature extractor and the TTSS features extracted from the same (there will be one \emph{TTSS} feature at each frequency) were then passed as inputs to the ML classifier, LR. This scheme, thus exploited the neurochaos based hybrid learning architecture for our task. Such an architecture has been shown to learn from finite sample datasets and can help to use chaos as a kernel trick which can be useful to make given data linearly separable.

\section{Results}
\label{sec_res}
The results obtained by using each of the methods discussed above are detailed in the subsections below. We term each of our methods as models, as what the methods are essentially trying to do is distinguish between a causal and a non-causal underlying structure. So, even though we are not strictly trying to fit a model, we are trying to learn a generalized characteristic of the temporal order in given data.

Each of the models were trained and tested using the {\bf AR training and testing set} which consisted of AR series as the causal time-series and time series with independent entries, randomly chosen from normal distribution, $\mathcal{N}(0,0.01)$, as the non-causal time series. We term time-series generated in the latter way as random time-series. 1250 AR and 1250 random time-series, each having a length of 2000 time points were simulated. Each AR series was of the form, $X(t)=a_k X(t-k) + \varepsilon_t$, with the order $k$, being randomly chosen between 1 and 20. These series were initialized to random values. The noise term followed the distribution $\mathcal{N}(0,0.01)$ and the AR coefficient for each simulation was randomly chosen such that $a_k \in U(0.8, 0.9)$. The training to testing split for this dataset was 70:30.

For further testing of the models, the following datasets were generated. These testing sets had a shift in the probability distribution of time-series when compared to the training set. We call them {\bf Distribution shift testing set I} and {\bf Distribution shift testing set II}. For both of these, causal time-series were generated in exactly the same manner as for the AR training and testing set. The non-causal time series in Distribution shift testing set I were generated using $\mathcal{N}(0,0.09)$ and in Distribution shift testing set II were generated using $U(-0.6,0.6)$. The number of non-causal time series in both datasets were 1250 and each time-series was simulated with 2000 time points. For these datasets, models were trained using the AR testing set described in the previous paragraph and tested based on these datasets. 

\subsection{Time-series values model}
\label{subsec_res_time_val_model}
Performance metrics for the time-series values model using both LR and LSTM on the simulated training and testing sets are shown in Table~\ref{table_time_val_res}. The non-causal time series were labelled as Class-0 and the causal time series as Class-1. The Precision, Recall and F1-Score columns are given in the format $(\cdot,\cdot)$, with the first value being for Class-0 and the second value for Class-1.

LR was implemented using the \emph{Scikit-Learn}~\cite{pedregosa2011scikit} package and LSTM was implemented using the \emph{Keras}~\cite{chollet2015keras} framework in Python. For LR, all parameters were kept as default. For LSTM, the layers were stacked in the following order: LSTM layer with 10 outputs, dropout layer with a parameter value of 0.5, dense hidden layer with 10 outputs and relu activation function and finally, a dense output layer with two outputs and soft-max activation function. This classification model used categorical cross-entropy as the loss function and
adam optimizer for optimization. The optimizer was run for 50 epochs
with a batch size of 1250. This classification model was run 15 times to avoid local minima and the run which gave the best testing accuracy was used.
\begin{table}[!h]
\caption{Prediction for time-series values model. LR: Logistic Regression, LSTM: Long Short-Term Memory. Scores are given in the format $(\cdot,\cdot)$, with the first value being for Class-0 and the second value for Class-1.}
\label{table_time_val_res}
\centering
\renewcommand{\arraystretch}{1.4}
\newcolumntype{C}[1]{>{\centering\arraybackslash}m{#1}}
\begin{tabular}{|C{3.1cm}|C{1.67cm}|C{1.77cm}|C{1.77cm}|C{1.77cm}|C{1.72cm}|}\hline
{\bf Dataset} & {\bf Classifier} & {\bf Precision} & {\bf Recall} & {$\mathbf{F_1 Score}$} & {\bf Accuracy}\\ \hline
\multirow{2}*{\bf AR training set} & {\bf LR} & (1.00, 1.00) & (1.00, 1.00) & (1.00, 1.00) & 100\%   \\ \cline{2-6}
& {\bf LSTM} & (0.98, 1.00) & (1.00, 0.98) & (0.99, 0.99) & 99\%   \\ \hline
\multirow{2}*{\bf AR testing set} & {\bf LR} & (0.53, 0.54) & (0.70, 0.37) & (0.60, 0.44) & 54\%   \\ \cline{2-6}
& {\bf LSTM} & (0.98, 0.99) & (0.99, 0.98) & (0.99, 0.99) & 99\%  \\ \hline
{\bf Distribution shift} & {\bf LR} & (0.49, 0.48) & (0.58, 0.38) & (0.53, 0.43) & 48\%   \\ \cline{2-6}
{\bf testing set I} & {\bf LSTM} & (0.15, 0.49) & (0.01, 0.97) & (0.01, 0.65) & 49\%  \\ \hline
{\bf Distribution shift} & {\bf LR} & (0.48, 0.47) & (0.56, 0.38) & (0.52, 0.42) & 47\%   \\ \cline{2-6}
{\bf testing set II} & {\bf LSTM} & (0.00, 0.49) & (0.00, 0.97) & (0.00, 0.65) & 48\%  \\
\hline
\end{tabular} \\
\end{table}

It can be seen from Table~\ref{table_time_val_res} that the LR classifier works well for the AR training set but fails for the AR testing set, indicating that it is overfitting the training set. Tuning the hyperparameters for the classifier did not help to improve its performance. LSTM gives good performance for both AR training as well as testing set but fails when there is a distribution shift in the random time-series. This is probably because the LSTM is recognizing only the random time-series used for training as non-causal and anything other than that as causal. Thus, it fails to learn the causal structure in time-series. 

\subsection{Frequency-domain representation model}

Performance metrics for the frequency-domain representation model using both LR and LSTM are shown in Table~\ref{table_fourier_res} The Precision, Recall and F1-Score columns are given in the format $(\cdot,\cdot)$, with the first value being for Class-0 (non-causal time series) and the second value for Class-1 (causal time series). In order to observe the characteristics of frequency amplitude of the signals, it would be essential to demean the time-series so as to remove any DC component that is present. However, since all our simulated datasets have a zero mean, we skipped this step.

The hyperparameters and architecture used for LR and LSTM as well as the packages used for the implementation  remained the same as in Section~\ref{subsec_res_time_val_model}.

\begin{table}[!h]
\caption{Prediction for frequency domain representation model. LR: Logistic Regression, LSTM: Long Short-Term Memory. Scores are given in the format $(\cdot,\cdot)$, with the first value being for Class-0 and the second value for Class-1.}
\label{table_fourier_res}
\centering
\renewcommand{\arraystretch}{1.4}
\newcolumntype{C}[1]{>{\centering\arraybackslash}m{#1}}
\begin{tabular}{|C{3.1cm}|C{1.7cm}|C{1.77cm}|C{1.77cm}|C{1.77cm}|C{1.62cm}|}\hline
{\bf Dataset} & {\bf Classifier} & {\bf Precision} & {\bf Recall} & {$\mathbf{F_1 Score}$} & {\bf Accuracy}\\ \hline
\multirow{2}*{\bf AR training set} & {\bf LR} & (0.99, 1.00) & (1.00, 0.99) & (1.00, 1.00) & 100\%   \\ \cline{2-6}
& {\bf LSTM} & (1.00, 1.00) & (1.00, 1.00) & (1.00, 1.00) & 100\%   \\ \hline
\multirow{2}*{\bf AR testing set} & {\bf LR} & (0.99, 1.00) & (1.00, 0.99) & (0.99, 0.99) & 99\%   \\ \cline{2-6}
& {\bf LSTM} & (1.00, 1.00) & (1.00, 1.00) & (1.00, 1.00) & 100\%  \\ \hline
{\bf Distribution shift} & {\bf LR} & (0.00, 0.50) & (0.00, 0.99) & (0.00, 0.66) & 50\%   \\ \cline{2-6}
{\bf testing set I} & {\bf LSTM} & (0.00, 0.50) & (0.00, 1.00) & (0.00, 0.67) & 50\%  \\ \hline
{\bf Distribution shift} & {\bf LR} & (0.00, 0.50) & (0.00, 0.99) & (0.00, 0.66) & 50\%   \\ \cline{2-6}
{\bf testing set II} & {\bf LSTM} & (0.00, 0.50) & (0.00, 1.00) & (0.00, 0.67) & 50\%  \\
\hline
\end{tabular} \\
\end{table}

\begin{figure}[h]
  \centering
 \includegraphics[width=0.35\linewidth]{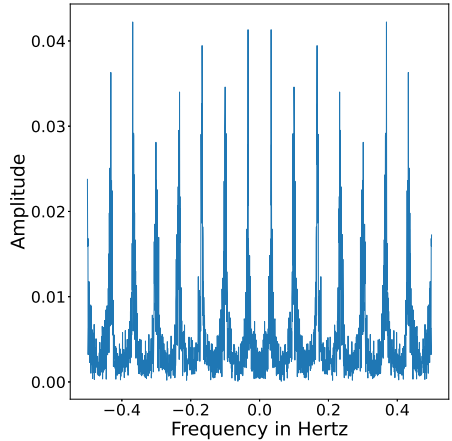}
  \caption{Frequency amplitude spectrum for a realization of AR(15) process}
  \label{fig_amp_causal}
\end{figure}

\begin{figure}[!h]
    \centering
    \includegraphics[scale=0.22]{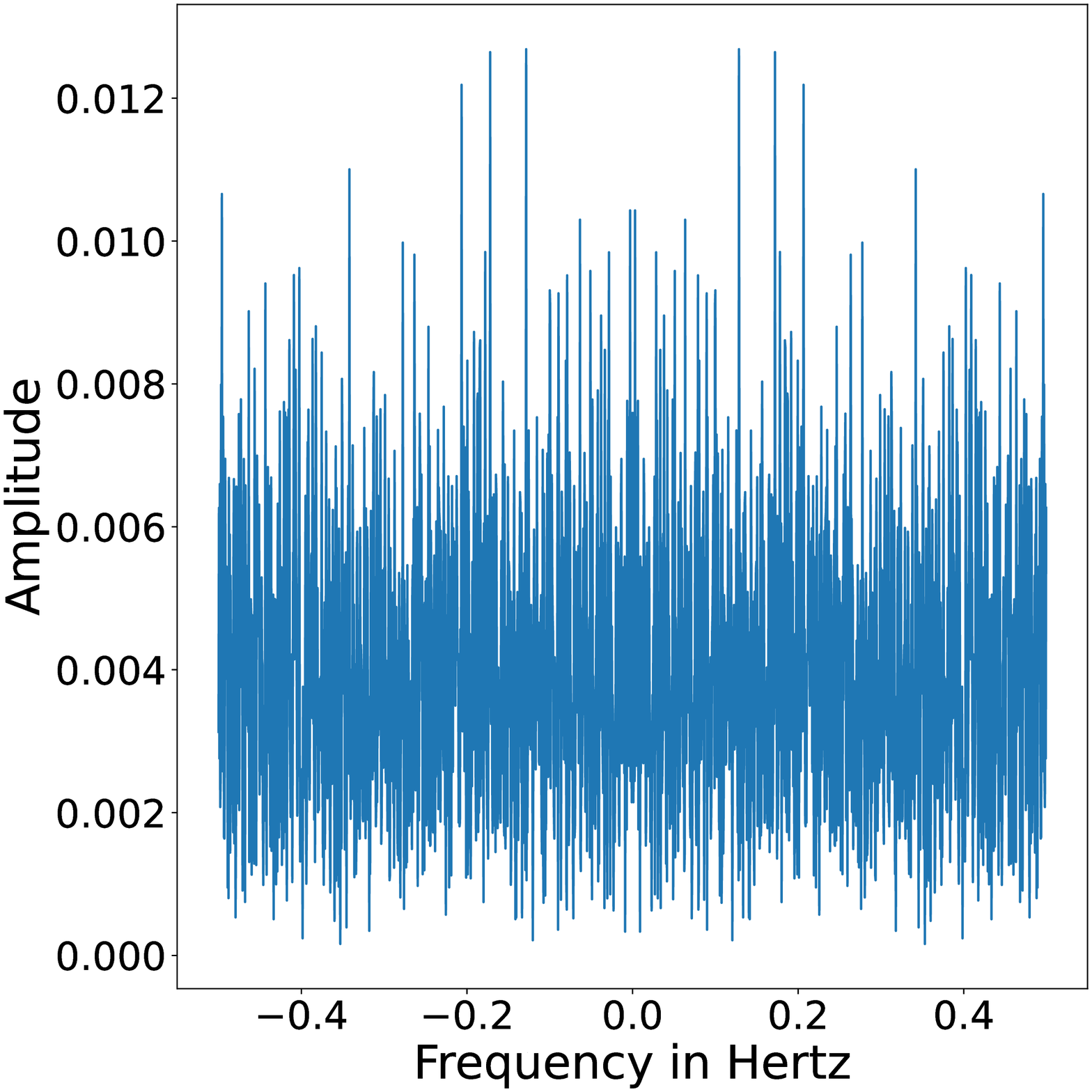}~~~
    \includegraphics[scale=0.22]{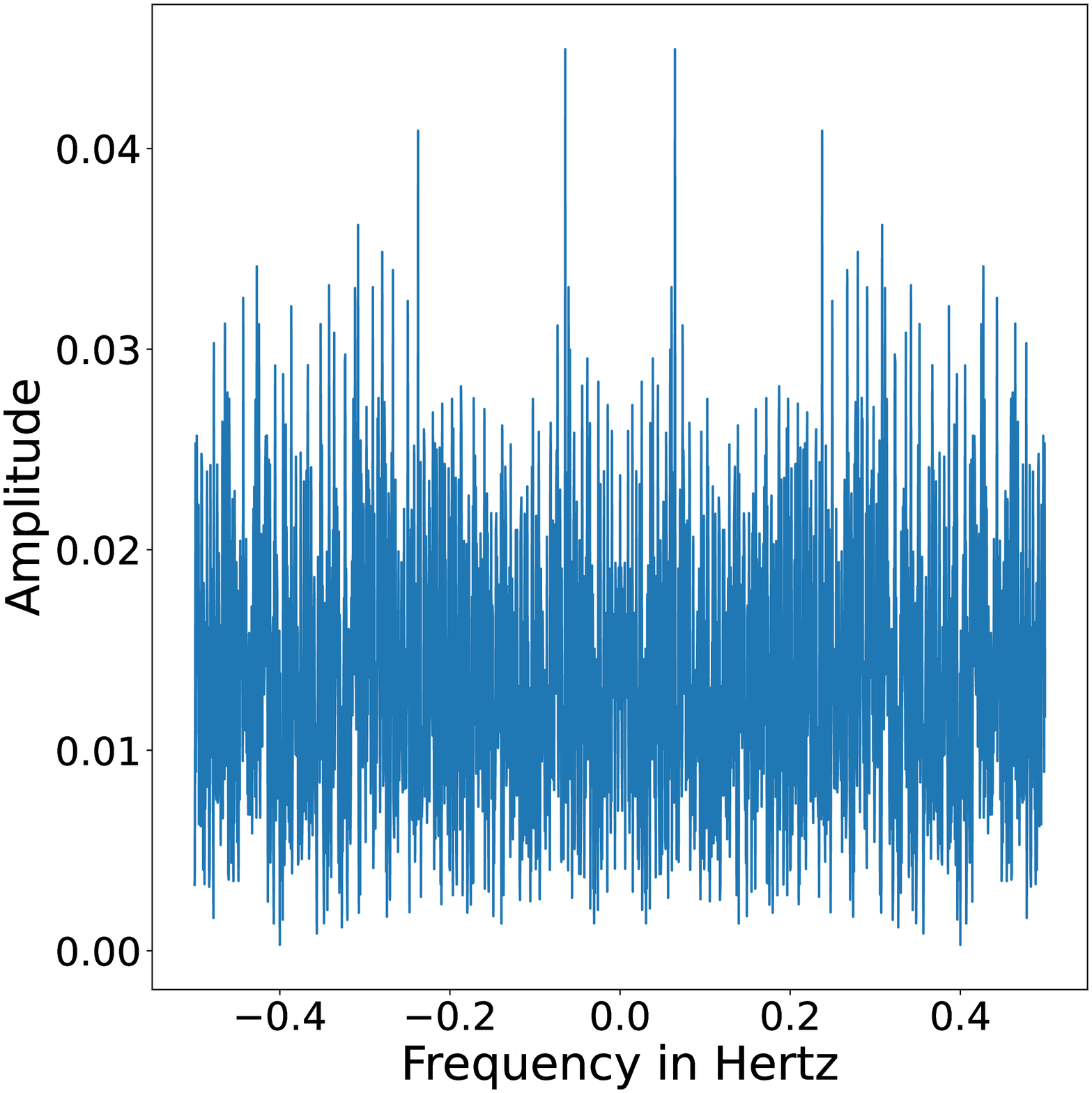}
    \begin{flushright}
(a)~~~~~~~~~~~~~~~~~~~~~~~~~~~~~~~~~~~~~~~~~~~~~~~~~~~~~~(b)~~~~~~~~~~~~~~~~~~~~~~~~~~~~~~~~~~~~~
\end{flushright}
    \caption{Frequency amplitude spectrum for a realization of random time-series generated using $\mathcal{N}(0,0.01)$ (left) and $U(-0.6,0.6)$ (right).}
    \label{fig_amp_noncausal}
\end{figure}

Figure~\ref{fig_amp_causal} shows the amplitude spectrum of the fourier transformed signal for a realization of AR(15) process. Figures~\ref{fig_amp_noncausal}(a) and~\ref{fig_amp_noncausal}(b) show the same for realizations of random time-series generated using distributions $\mathcal{N}(0,0.01)$ and $U(-0.6,0.6)$ respectively. It can be seen from these figures that the frequency domain characteristics of causal and non-causal time series used are quite different, yet the implemented ML classifiers are unable to distinguish between the two.


It can be seen from Table~\ref{table_fourier_res} that both LR and LSTM accurately classify the AR training and testing sets, however, fail for testing sets with distribution shifts in non-causal time-series. This again seems to be because the classifiers are recognizing only the random time-series used for training as non-causal and anything other than that as causal and have failed to learn any causal-structure from the data.

\subsection{ChaosFEX feature representation model (FT+ChaosFEX)}

Performance metrics when TTSS features of the amplitudes at different frequencies were passed as inputs to the ML classifier logistic regression are shown in Table~\ref{table_cfex_res}. The Precision, Recall and F1-Score columns are given in the format $(\cdot,\cdot)$, with the first value being for Class-0 (non-causal time series) and the second value for Class-1 (causal time series).
\begin{table}[h]
\caption{Prediction for FT+ChaosFEX model with Logistic Regression classifier. Scores are given in the format $(\cdot,\cdot)$, with the first value being for Class-0 and the second value for Class-1.}
\label{table_cfex_res}
\centering
\renewcommand{\arraystretch}{1.4}
\newcolumntype{C}[1]{>{\centering\arraybackslash}m{#1}}
\begin{tabular}{|C{3.8cm}|C{1.77cm}|C{1.77cm}|C{1.77cm}|C{1.62cm}|}\hline
{\bf Dataset} & {\bf Precision} & {\bf Recall} & {$\mathbf{F_1 Score}$} & {\bf Accuracy}\\ \hline
{\bf AR training set} & (1.00, 1.00) & (1.00, 1.00) & (1.00, 1.00) & 100\%   \\ \hline
{\bf AR testing set} & (1.00, 1.00) & (1.00, 1.00) & (1.00, 1.00) & 100\%   \\ \hline
{\bf Distribution shift testing set I} & (1.00, 1.00) & (1.00, 1.00) & (1.00, 1.00) & 100\%   \\ \hline
{\bf Distribution shift testing set II} & (1.00, 1.00) & (1.00, 1.00) & (1.00, 1.00) & 100\%   \\ 
\hline
{\bf AR(100) testing set} & (NA, 1.00) & (NA, 0.99) & (NA, 1.00) &99\%   \\ \hline
{\bf ARMA testing set} & (NA, 1.00) & (NA, 1.00) & (NA, 1.00) & 100\%  \\ \hline
{\bf ARFIMA testing set} & (NA, 1.00) & (NA, 1.00) & (NA, 1.00) & 100\%   \\ \hline
\end{tabular} \\
\end{table}

It can be seen that this model classifies accurately not just the AR training and testing sets but also both the testing sets with distribution shifts in the non-causal time series. To further check the robustness of the model, it was tested on more testing sets in which there was a shift in the distribution of causal time-series. Three testing sets generated in this manner included AR(100), ARMA and ARFIMA as causal time-series and no non-causal time series. Since these datasets did not include any non-causal time series, the first value for precision, recall and $F_1$ score are marked as NA (Not-Applicable) in Table~\ref{table_cfex_res}. 

The specific details of the three new testing sets mentioned above are as follows. Each of these datasets consisted of 1250 causal time-series with 2000 time points each. Each time series was initialized to random values and had an instantaneous noise term, $\varepsilon_t \in \mathcal{N}(0,0.01)$. AR(100) processes followed the form $X(t)=a_{100} X(t-100) + \varepsilon_t$, where, for each realization, $a_{100} \in U(0.8, 0.9)$. For each realization in the set of ARMA and ARFIMA processes, the AR and MA orders were randomly chosen between 1 and 20 and the AR and MA coefficients were randomly chosen from $U(0.8,0.9)$. The difference parameter $d$ was randomly chosen from $U(-0.5, 0.5)$ for ARFIMA processes.

It can be seen from the performance metrics in Table~\ref{table_cfex_res} that the model worked extremely well for accurate classification of above discussed causal time-series with distribution shifts and with the presence of causal influence that was occurring at very different scales as compared to the influence in the causal-time series used in the training set.      

\begin{figure}[h!]
    \centering
    \includegraphics[scale=0.22]{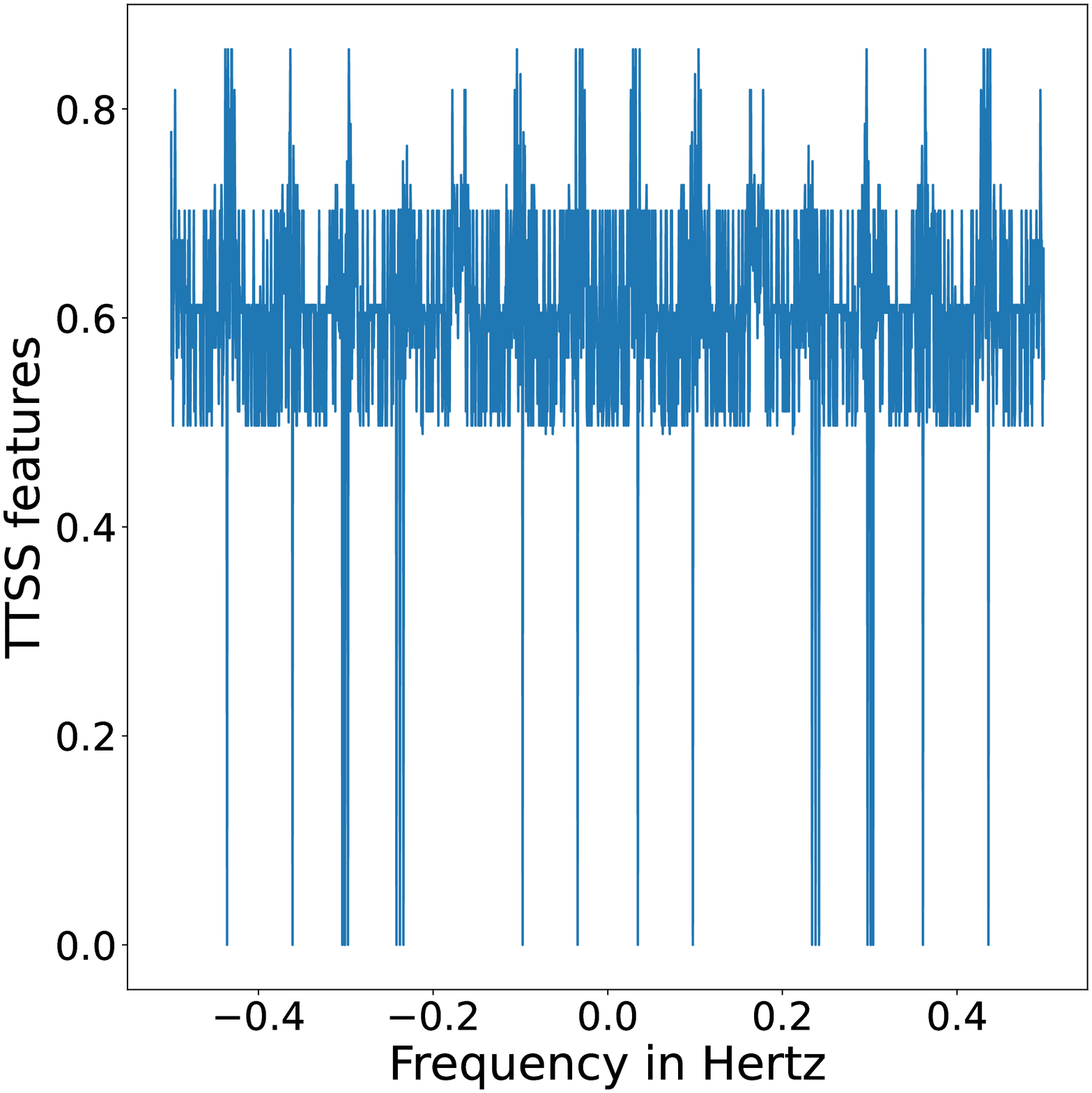}~~~
    \includegraphics[scale=0.22]{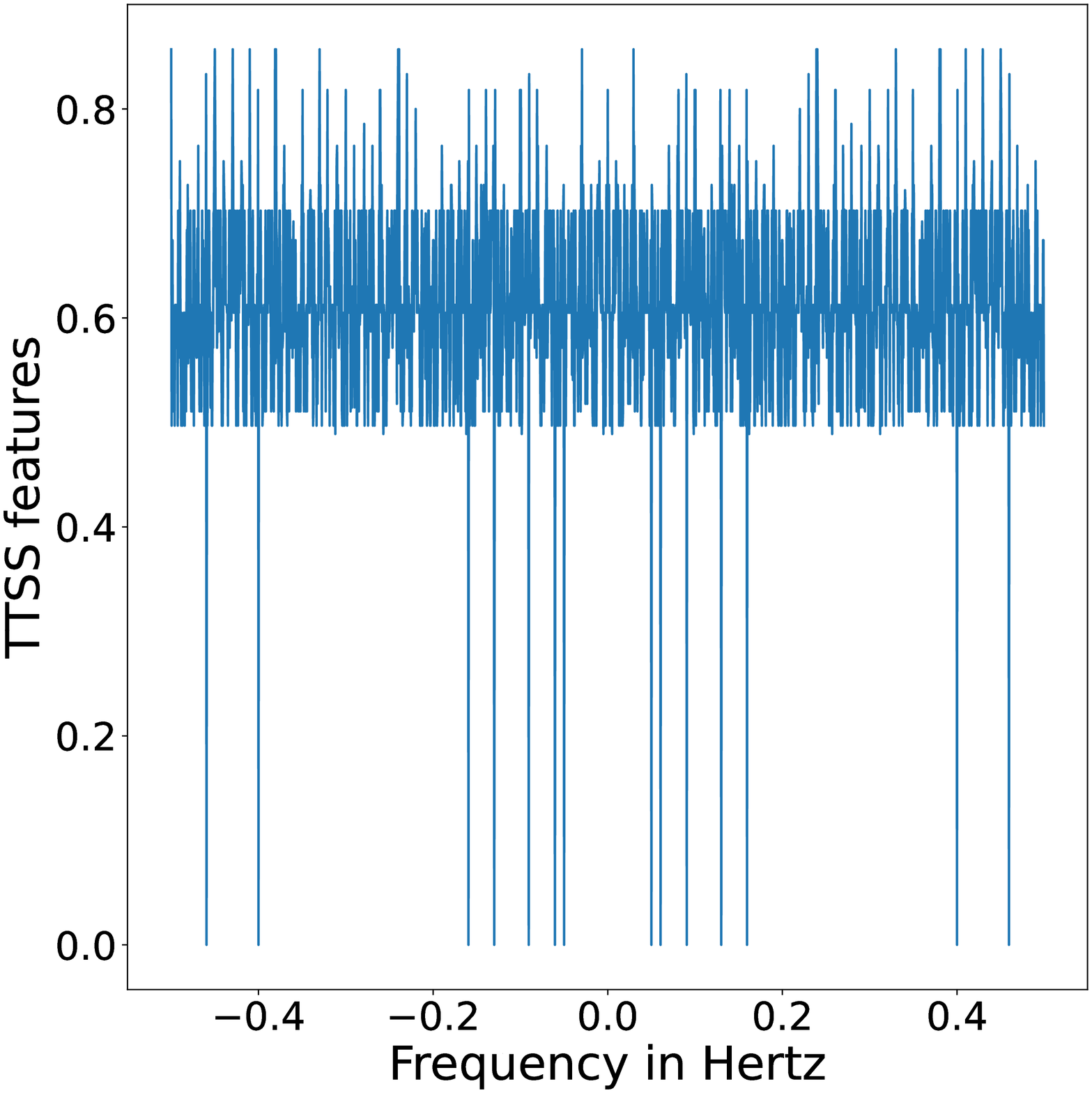}
    \begin{flushright}
(a)~~~~~~~~~~~~~~~~~~~~~~~~~~~~~~~~~~~~~~~~~~~~~~~~~~~~~~(b)~~~~~~~~~~~~~~~~~~~~~~~~~~~~~~~~~~~~~
\end{flushright}
\includegraphics[scale=0.22]{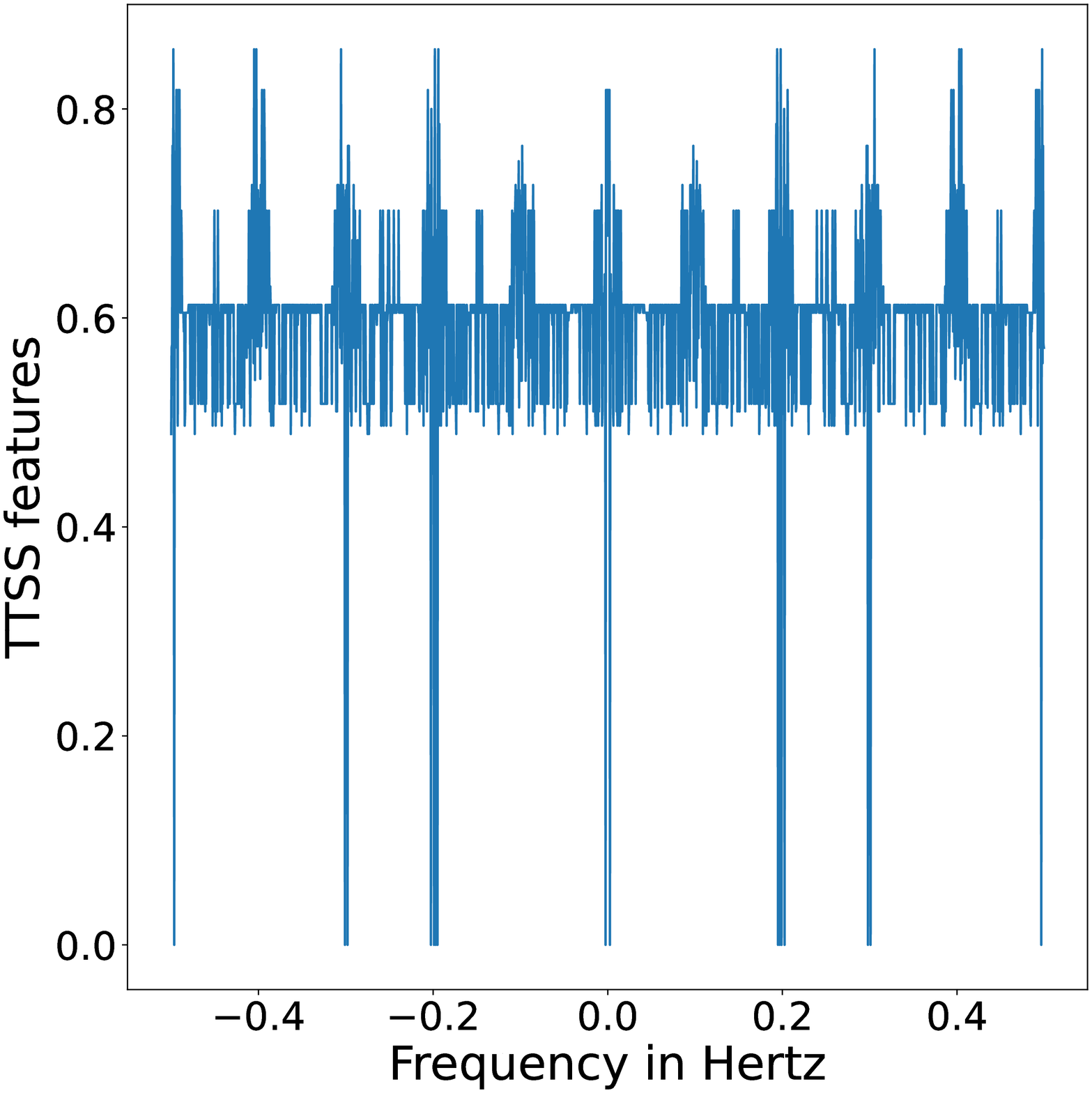}~~~
    \includegraphics[scale=0.22]{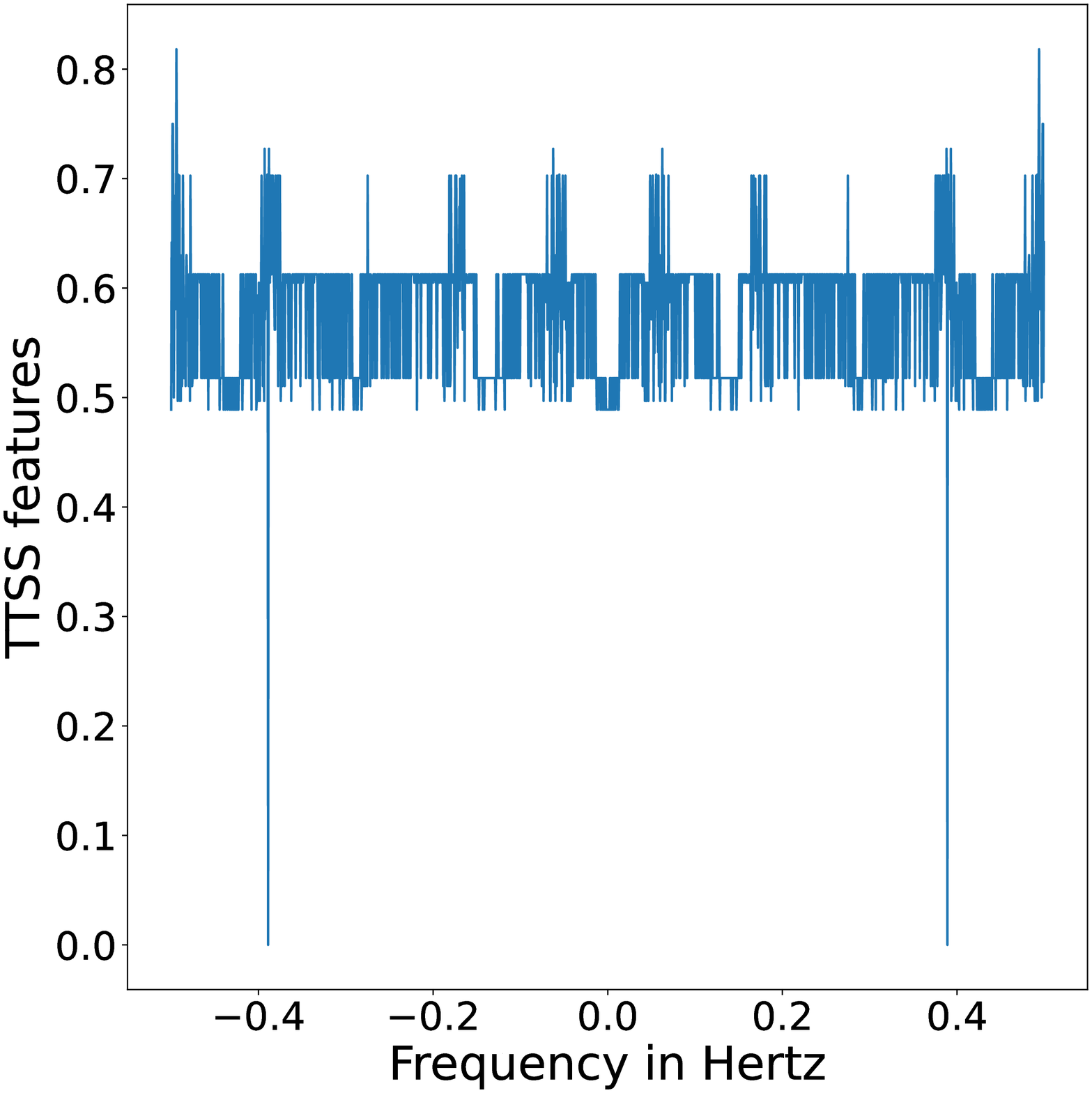}
    \begin{flushright}
(c)~~~~~~~~~~~~~~~~~~~~~~~~~~~~~~~~~~~~~~~~~~~~~~~~~~~~~~(d)~~~~~~~~~~~~~~~~~~~~~~~~~~~~~~~~~~~~~
\end{flushright}
    \caption{TTSS feature representation for the amplitude at each frequency for causal time-series: (a)AR(15), (b)AR(100), (c)ARMA and (d)ARFIMA}
    \label{fig_TTSS_causal}
\end{figure}

\begin{figure}[h]
    \centering
    \includegraphics[scale=0.22]{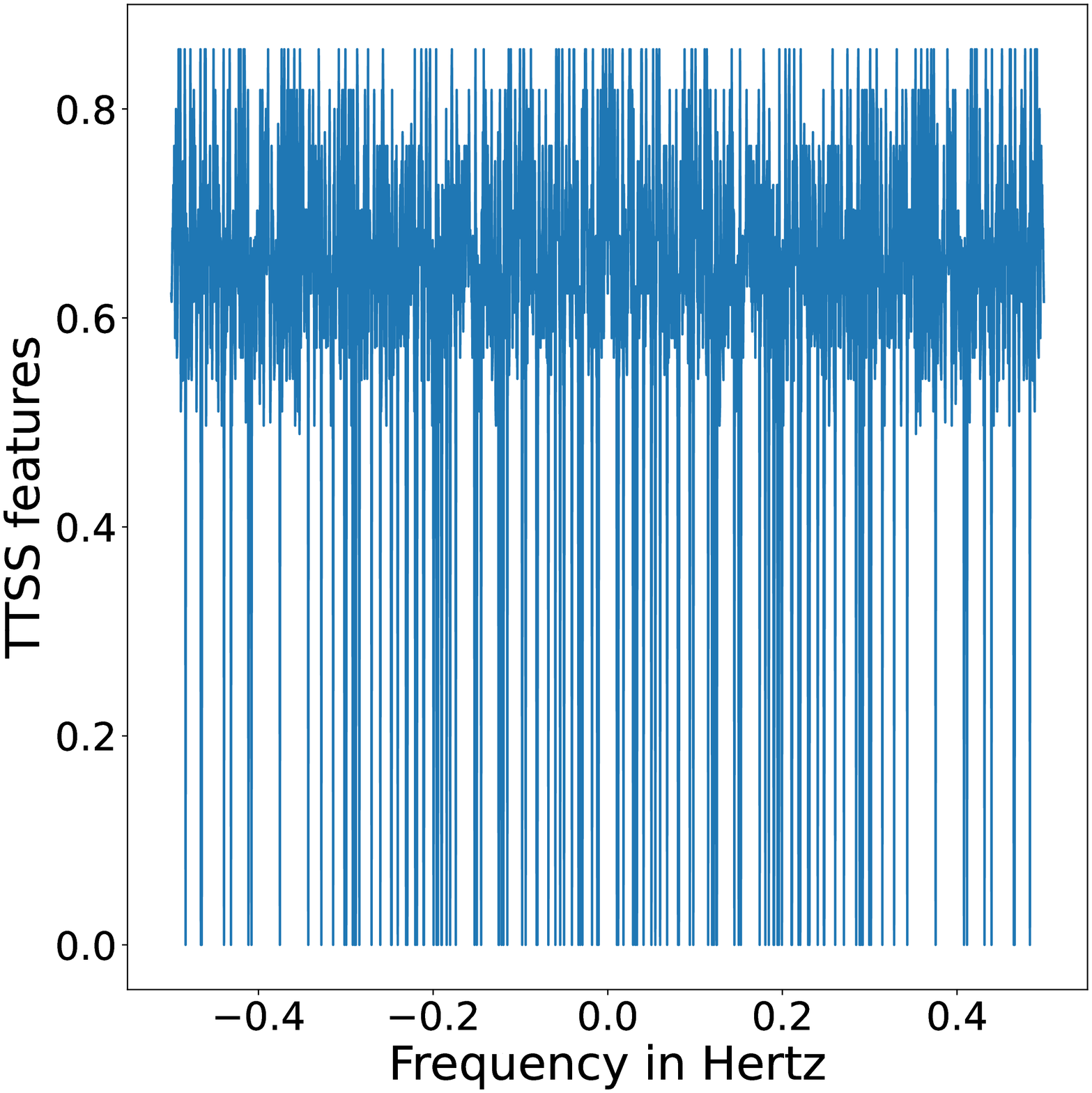}~~~
    \includegraphics[scale=0.22]{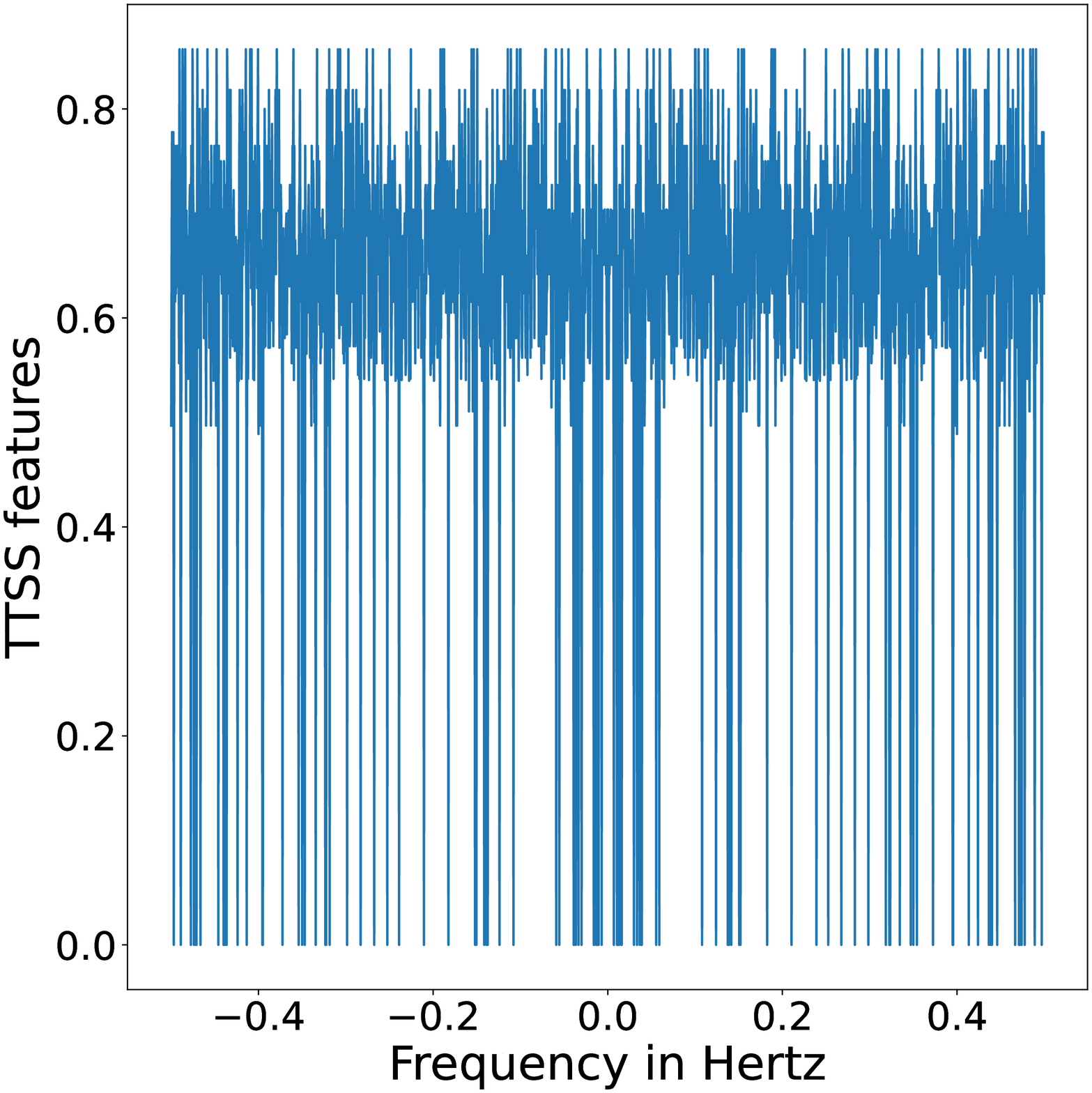}
    \begin{flushright}
(a)~~~~~~~~~~~~~~~~~~~~~~~~~~~~~~~~~~~~~~~~~~~~~~~~~~~~~~(b)~~~~~~~~~~~~~~~~~~~~~~~~~~~~~~~~~~~~~
\end{flushright}
    \caption{TTSS feature representation for the amplitude at each frequency for non-causal time-series generated from: (a) $\mathcal{N}(0,0.01)$ and (b) $U(-0.6,0.6)$.}
    \label{fig_TTSS_noncausal}
\end{figure}

We also plot figures to illustrate the difference in TTSS features of the amplitudes at different frequencies for causal and non-causal time series used in this section. Figures~\ref{fig_TTSS_causal}(a),~\ref{fig_TTSS_causal}(b),~\ref{fig_TTSS_causal}(c) and~\ref{fig_TTSS_causal}(d) show the figures for causal time series, AR(15), AR(100), ARMA and ARFIMA respectively. Figures~\ref{fig_TTSS_noncausal}(a) and~\ref{fig_TTSS_noncausal}(b) show the figures for non-causal time series generated using $\mathcal{N}(0,0.01)$ and $U(-0.6,0.6)$ respectively. Clearly the number of peaks and valleys in TTSS feature plots of causal time-series are much lower than those in TTSS feature plots of non-causal time series.

Hyperparameters used for ChaosFEX feature extraction for all datasets in this section are $q=0.33$, $b=0.499$, $\varepsilon=0.01$ and maximum trajectory length of firing for each GLS neuron was set to 1000. This means that the GLS neuron would stop firing after a maximum trajectory length of 1000 if it did not reach the $\varepsilon$ neighborhood of the stimulus. In~\cite{nb2020neurochaos} and~\cite{harikrishnan2021noise}, the authors have found the hyperparameters tuned to $q=0.34$, $b=0.499$ and $\varepsilon$ in the range $0.005$ to $0.3$, to be the most effective for the classification of sequential data such as genome sequences and spoken digits. Thus we fine tuned the hyperparameters by limiting the exploration of the parameters around this range. Hyperparameter tuning for the LR classifier was also done in this case. All parameters were set to default values other than \emph{`$max\_iter$', `tol'} and \emph{`C'}, which were set to $1000, 0.001$ and $0.001$ respectively.

\section{Discussion, concluding remarks and future research directions}
\label{sec_discussion}
We show that ML classifiers (Logistic regression and LSTM), when used by themselves directly on time-series measurements are dumb to the temporal/ causal-structure in the data. This fact has also been discussed in existing literature~\cite{pearl2018book, scholkopf2019causality}. When time-series values were directly passed to the classifiers, LR and LSTM, they failed to learn any causal-structure characteristics from the data. Even though LSTM has been developed for the classification of sequential data, it seemed to be learning some statistical features from the data, giving high classification accuracy when the testing set followed the same distribution as the training set and failing when there was a distribution shift in the non-causal time series structure.

Even though characteristics of the fourier transformed time series of causal and non-causal type are quite different, classifiers LR and LSTM, when directly used on the frequency domain amplitudes fail to learn the causal structure. Though they could accurately classify testing set which was exactly the same as training set, the classification accuracy dropped to about $50\%$, when there was a change in the distribution of non-causal time series. Almost all of these time series were being wrongly classified as causal. 

TTSS feature representation of the amplitude values at different frequencies proved to be robust in learning the causal structure. We show that it works in case of distribution shifts in both causal and non-causal time series and could learn a generalized structure, identifying the processes correctly irrespective of the scale at which causal-influence existed in the processes. Good performance of the developed pipeline on our simple problem illustrates the power of chaos and of `neurochaos' based hybrid learning architectures for developing more sophisticated causality based ML architectures as well as in using ML for causal inference tasks. Since, even a simple linear classifier, LR, is able to distinguish between TTSS features from causal and non-causal time series, the strength of this feature seems to lie in the ability to transform the given data to make it linearly separable. Hence, the developed pipeline seems to be helpful in strong feature extraction for causal-structure learning purposes.

Future work would involve more rigorous testing of the TTSS based pipeline with other datasets having different levels of noise, strength of causal coefficient and different causal/ non-causal structures. Simulated data with `interventional perturbations' will also be provided to the algorithms to test their performance on such data. Understanding and classification of these kind of cause-effect treatment datasets will benefit the most from the application of the proposed approach. Using classifiers other than LR, for example Support Vector Machine with a linear kernel, to classify based on the TTSS model will also be done. Other features based on the chaotic firing trajectory of GLS neurons in the Chaosnet architecture have also been utilized in hybrid ML based learning tasks~\cite{harikrishnan2020neurochaos, nb2020neurochaos}. We would also like to check the performance of the ChaosFEX based model using features other than the TTSS feature (or GLS firing rate) for classification of causal and non-causal time-series. Experiments and analysis will be done to get a better theoretical understanding of how and why the ChaosFEX model works well for causal-structure learning tasks. Finally, we would like to apply the developed technique to recognize real time-series data with causal structure. We would like to check the performance of the method for cause-effect treatment estimation tasks for real data, both when the treatment/intervention being provided is a discrete event or a continuous event. There are still not many well developed techniques available for the latter purpose~\cite{moraffah2021causal} and the proposed pipeline seems like a promising approach. 

\section*{Acknowledgment}
The authors are thankful to Harikrishnan N.B., National Institute of Advanced Studies, for providing help with the use of ChaosFEX toolbox. N. Nagaraj gratefully acknowledges the financial support of Tata Trusts and Dept. of Science and Tech., Govt. of India (grant no. DST/CSRI/2017/54). A. Kathpalia acknowledges the financial support of the Czech Science Foundation, Project No. GA19-16066S and the Czech Academy of Sciences, Praemium Academiae awarded to M. Paluš.
%
\bibliographystyle{unsrt}  
{
\small
\bibliography{references} 
}

%
%
%
%
\medskip


\end{document}